\documentclass[12pt]{article}
\usepackage{amssymb}
\usepackage{amsfonts,amsmath}

\usepackage{xspace}                           
\usepackage{booktabs}                         
\usepackage{hyperref}                         
\usepackage{graphicx}

\newtheorem{theorem}             {Theorem}
\newtheorem{lemma}      [theorem]{Lemma}
\newtheorem{corollary}  [theorem]{Corollary}

\newtheorem{definition} [theorem]{Definition}
\newtheorem{proposition}[theorem]{Proposition}

\newcommand{\expect}[1]{\mathbf{E}\left[#1\right]}

\def\N{\mathbb{N}}
\def\x{{x}}
\def\y{{y}}

\newcommand{\psel}{\ensuremath{p_\text{sel}}\xspace}
\newcommand{\pmut}{\ensuremath{p_\text{mut}}\xspace}
\newcommand{\pxor}{\ensuremath{p_\text{xor}}\xspace}

\DeclareMathOperator{\unif}{Unif}

\usepackage{mathrsfs}

\usepackage{algorithm,algorithmic}
\floatname{algorithm}{Algorithm}

\usepackage{enumitem}

\usepackage{color}

\newcommand{\qed}{\hspace*{\fill} \rule{1ex}{1ex} \vspace{0.5ex}}

\begin{document}
\pagestyle{empty}

\title{Level-based Analysis of Genetic Algorithms for Combinatorial Optimization}

\author{Duc-Cuong Dang, Anton V. Eremeev, Per Kristian Lehre}

\maketitle

\begin{abstract}
The paper is devoted to upper bounds on run-time of Non-Elitist
Genetic Algorithms until some target subset of solutions is
visited for the first time. In particular, we consider the sets of
optimal solutions and the sets of local optima as the target
subsets. Previously known upper bounds are improved by means of
drift analysis. Finally, we propose conditions ensuring that a
Non-Elitist Genetic Algorithm efficiently finds approximate
solutions with constant approximation ratio on the class of
combinatorial optimization problems with guaranteed local
optima~(GLO).
\end{abstract}


\section*{\large{\textbf{Introduction}}}

The genetic algorithm (GA) proposed by J.~Holland~\cite{Holl75} is
a randomized heuristic search method, based on analogy with the
genetic mechanisms observed in nature and employing a population
of tentative solutions. Different modifications of GA are widely
used in the areas of operations research, pattern recognition,
artificial intelligence etc. (see e.g.~\cite{Reeves,Yank99}).
Despite of numerous experimental investigations of these
algorithms, their theoretical analysis is still at an early
stage~\cite{BSW}.

Efficiency of a GA in application to a combinatorial optimization
problem may be estimated in terms of expected computation time
until an optimal solution or an acceptable approximation solution
is visited for the first time. It is very unlikely, however, that
there exists a randomized algorithm finding a globally optimal
solution for an NP-hard optimization problem on average in
polynomially bounded time. This would contradict the well-known
hypothesis ${\mbox{\rm NP} \neq \mbox{\rm RP}}$ which is in use
for several decades~\cite{Ko82}.


The main results of this paper are obtained through comparison of
genetic algorithms to local search, which is motivated by the fact
that the GAs are often considered to be good at finding local
optima (see e.g.~\cite{AartsLenstra,Krish,RR02}).

Here and below we assume that the randomness is generated only by
the randomized operators of selection, crossover, mutation and
random initialization of population within the GA (the input data
is deterministic). A function of input data is called {\em
polynomially bounded}, if there exists a polynomial in the length
of the problem input, which bounds the function from above. The
terms {\em efficient algorithm} or {\em polynomial-time algorithm}
are used for an algorithm with polynomially bonded running time.

\section{Combinatorial Optimization Problems and Genetic Algorithms}\label{sec:defs}

\paragraph{NP Optimization Problems}
In this paper, the combinatorial optimization problems are viewed
under the technical assumptions of the class of NP~optimization
problems (see e.g.~\cite{ACGKMP}). Let $\{0,1\}^*$ denote the set
of all strings with symbols from~$\{0,1\}$ and arbitrary string
length. For a string~$S\in \{0,1\}^*$, the symbol~$|S|$ will
denote its length. In what follows, $\N$ denotes the set of
positive integers and given a string~$S\in \{0,1\}^*$, the
symbol~$|S|$ denotes the length of the string~$S$. To denote the
set of polynomially bounded functions we define $\mbox{\rm Poly}$
as the class of functions from $\{0,1\}^*$ to $\N$ bounded above
by a polynomial in~$|I|,$ where $I \in \{0,1\}^*$.

\begin{definition}\label{def:NPO} An NP~optimization problem $\Pi$
is a triple ${\Pi=(\mbox{\rm Inst},\mbox{\rm Sol}(I),F_I)}$, where
$\mbox{\rm Inst} \subseteq \{0,1\}^*$ is the set of instances
of~$\Pi$ and:

1. The relation $I \in \mbox{\rm Inst}$ is computable in
polynomial time.

2. Given an instance $I \in \mbox{\rm Inst}$, $\mbox{\rm
Sol}(I)\subseteq \{0,1\}^{n(I)}$ is the set of feasible solutions
of~$I$, where~$n(I)$ stands for the dimension of the search
space~${\mathcal X}_I:=\{0,1\}^{n(I)}$. Given $I \in \mbox{\rm
Inst}$ and $\x \in \{0,1\}^{n(I)}$, the decision whether $\x\in
\mbox{\rm Sol}(I)$ may be done in polynomial time, and $n(\cdot)
\in \mbox{\rm Poly}$.

3. Given an instance~$I \in \mbox{\rm Inst}$,  $F_I: \mbox{\rm
Sol}(I) \to {\N}$ is the objective function (computable in
polynomial time) to be maximized if $\ \Pi$ is an NP~maximization
problem or to be minimized if $\ \Pi$ is an NP~minimization
problem.
\end{definition}

Without loss of generality we will consider only the maximization
problems. The results will hold for the minimization problems as
well. The symbol of problem instance~$I$ may often be skipped in
the notation, when it is clear what instance~$I$ is meant.

\begin{definition}\label{PolyBoundedProblem}
A combinatorial optimization problem $\Pi=(\mbox{\rm
Inst},\mbox{\rm Sol}(I),F_I)$ is {\em polynomially bounded}, if
there exists a polynomial in~$|I|$, which bounds the objective
values $F_I({x})$, ${x \in {\rm Sol}(I)}$ from above.
\end{definition}

\paragraph{Neighborhoods and local optima}
Let a neighborhood ${\mathcal N}_I({y})\subseteq \mbox{\rm
Sol}(I)$ be defined for every~${y}\in \mbox{\rm Sol}(I)$. The
mapping ${\mathcal N}_I: {\rm Sol}(I) \to 2^{{\rm Sol}(I)}$ is
called the {\em neighborhood mapping}. Following~\cite{AP95}, we
assume this mapping to be efficiently computable, i.e. the set
${\mathcal N}_I(x)$ may be enumerated in polynomial time.

\begin{definition} If the inequality $F_I({y})\leq F_I({x})$
holds for all neighbors~$\y \in {\mathcal N}_I({x})$ of a
solution~${{x} \in \mbox{\rm Sol}(I)}$, then~${x}$ is called a
local optimum w.r.t. the neighborhood mapping~${\mathcal N}_I$.
\end{definition}

Suppose $R(\cdot,\cdot)$ is a metric on $\mbox{\rm Sol}(I)$. The
neighborhood mapping
$$
{\mathcal N}_I({x})=\{{y} \mid R({x},{y}) \leq r\}, \ \ {x} \in
\mbox{\rm Sol}(I),
$$
is called a neighborhood mapping of radius~$r$ defined by
metric~$R(\cdot,\cdot)$.


A local search method starts from some feasible solution~$\y_0$.
Each iteration of the algorithm consists in moving from the
current solution to a new solution in its neighborhood, such that
the value of objective function is increased. The way to choose an
improving neighbor, if there are several of them, will not matter
in this paper. The algorithm continues until a local optimum is
reached.

\paragraph{Genetic Algorithms}
The Simple GA proposed in~\cite{Holl75} has been intensively
studied and exploited over four decades. A plenty of variants of
GA have been developed since publication of the Simple GA, sharing
the basic ideas, but using different population management
strategies, selection, crossover and mutation
operators~\cite{RR02}.

The GA operates with
populations~$P^t=({x}^{1,t},\dots,{x}^{\lambda, t})$, \
$t=0,1,\dots,$ which consist of~$\lambda$ {\it genotypes}. In
terms of the present paper the genotypes are elements of the
search space~$\mathcal X$.

In a {\em selection} operator~${\mbox{Sel}: {\mathcal X}^{\lambda}
\to \{1,\dots,\lambda\}}$, each parent is independently drawn from
the previous population~$P^t$ where each individual in~$P^t$ is
assigned a selection probability depending on its {\em
fitness}~$f({x})$. Usually a higher fitness value of an individual
implies higher (or equal) selection probability. Below we assume
the following natural form of the fitness function:

\begin{itemize}

\item if $\x\in {\rm Sol}$ then
$$f({x}) = F({x});$$

\item if ${x}\not \in {\rm Sol}$ then its fitness is
defined by some penalty function, such that
$$
f({x}) < \min_{\y \in {\rm Sol}} F(\y).
$$
\end{itemize}

In this paper, we consider the tournament selection,
$(\mu,\lambda)$-selection
and exponential ranking selection
(see the details in Section~\ref{sec:beta_lower_bounds} below).

One or two offspring genotypes is created from two parents using
the randomized operators of crossover~$\mbox{Cross}: {\mathcal X}
\times {\mathcal X} \to {\mathcal X} \times {\mathcal X}$
(two-offspring version) or~${\mbox{Cross}}: {\mathcal X} \times
{\mathcal X} \to {\mathcal X}$ (single-offspring version) and
mutation ${\mbox{Mut}:{\mathcal X} \to {\mathcal X}}$. In general,
we assume that $\mbox{Cross}(\x,\y)$ and $\mbox{Mut}(\x)$ are
efficiently computable randomized routines.

When a population~$P^{t}$ of~$\lambda$ offspring is constructed,
the GA proceeds to the next iteration~$t+1$. An initial
population~$P^{0}$ is generated randomly. One of the ways of
initialization consists, e.g. in independent choice of all bits in
genotypes.

To simplify the notation below, $\mathcal GA$ will always denote
the non-elitist genetic algorithm with single-offspring crossover
based on the following outline.\pagebreak

{\bf Algorithm $\mathcal GA$}\\
\vspace{-0.5em}

\noindent Generate the initial population~$P^0$, assign $t:=1.$\\
 {\bf While} termination condition is not met {\bf do:}\\
$\mbox{\hspace{2em}}$ {\bf Iteration $t$.}\\
$\mbox{\hspace{2em}}$ {\bf For $j$ from 1 to ${\lambda}$ do:} \\
$\mbox{\hspace{4em}}$   Selection: $i:={\rm
Sel}(P^{t-1})$, $i':={\rm Sel}(P^{t-1})$. \\
$\mbox{\hspace{4em}}$   Crossover: $\x:=\mbox{Cross}(\x^{i,t-1},\x^{i',t-1}).$\\
$\mbox{\hspace{4em}}$   Mutation: $\x^{j,t} := \mbox{Mut}(\x).$\\
$\mbox{\hspace{2em}}$ {\bf End for.}\\
$\mbox{\hspace{2em}}$ $t:=t+1.$\\
 {\bf End while.}\\


In theoretical analysis of the $\mathcal GA$ we will assume that
the termination condition is never met. The termination condition,
however, may be required to stop a genetic algorithm when a
solution of sufficient quality is obtained or the computing time
is limited, or because the population is "trapped" in some
unpromising area and it is preferable to restart the search (see
e.g.~\cite{BN98,Vitanyi}).

%
%

In what follows, the operators of selection, mutation and
single-offspring crossover are associated with the corresponding
transition matrices:
\begin{itemize}
  \item $\psel:[\lambda]\to [0, 1]$
represents a selection operator, where 
$\psel(i|P_t)$ is the probability of selecting the~$i$-th
individual from population~$P_t$.
 \item $\pmut :
{\mathcal X} \times {\mathcal X} \to [0, 1]$, where~$\pmut(y|x)$
is the probability of mutating $x\in {\mathcal X}$ into~$y \in
{\mathcal X}$.
 \item $\pxor \colon {\mathcal X} \times {\mathcal X}^2
\to [0, 1]$, where~$\pxor(x'|x,y)$ is the probability of
obtaining~$x'$ as a result of crossover between~$x,y \in {\mathcal
X}$.
\end{itemize}

The single-offspring crossover may be obtained from two-offspring
crossover by first computing $(u,v):=\mbox{\rm Cross}(x,y)$, and
then defining $x':=\mbox{\rm Cross}(x,y)$ as $x' \sim
\unif(\{u,v\})$.

\paragraph{Crossover and Mutation Operators}
Let us consider the well-known operators of bitwise
mutation~$\mbox{Mut}^*$ and the single-point
crossover~$\mbox{Cross}^*$ from Simple GA~\cite{bib:Gold89} as
examples.

The single-point crossover operator computes
$({x}',{y}')=\mbox{Cross}^*({x},{y})$, given $ {x}=(x_1,...,x_n
),$ ${y}=(y_1,..., y_n),$ so that with a given probability~$p_{\rm
c}$,
$$
{x}'=(x_1,...,x_{Z}, y_{Z+1},...,y_n ), \ \ {y}'=(y_1,..., y_{Z},
x_{Z+1},..., x_n),
$$
where the random number~$Z$ is chosen uniformly from~1 to~$n-1$.
With probability ${1-p_{\rm c}}$ both parent individuals are
copied without any changes, i.e. ${x'=x}, \ {y'=y}$.

The bitwise mutation operator~$\mbox{Mut}^*$ computes a
genotype~${y}=\mbox{Mut}^*({x})$, where independently of other
bits, each bit~$y_i, \ i\in [n]$, is assigned a value~$1-x_i$ with
probability~$p_{\rm m}$ and with probability~$1-p_{\rm m}$ it
keeps the value~$x_i$. Here and below we use the notation~$[n] :=
\{1, 2,...,n\}$ for any positive integer~$n$. The tunable
parameter~$p_{\rm m}$ is also called {\em mutation rate}.

The following condition holds for many well-known cross\-over
operators: there exists a positive constant~$\varepsilon_1$ which
does not depend on~$I$, such that given a pair of bitstrings~$x,
y$, the crossover result $x'=\mbox{\rm Cross}(x,y)$ satisfies
\begin{equation} \label{eqn:eps1_def}
\varepsilon_1 \le \left\{
\begin{array}{ll}
Pr\big(f(x')=f(x)\big), & \mbox{if} \ f(x)=f(y),\\
Pr\big(f(x')>\min\{f(x),f(y)\}\big), & \mbox{otherwise}.
\end{array} \right.
\end{equation}
This condition is fulfilled for the single-point crossover with
$\varepsilon_1 = 1-p_{\rm c}$, if $p_{\rm c}<1$ is a constant.
Sometimes stronger statements can be deduced, e.g. for the
well-known OneMax and LeadingOnes fitness functions the offspring
has a fitness $(f(x)+f(y))/2$ with probability at least~$1/2$
(see~\cite{CDEL14}).

Another condition analogous to~(\ref{eqn:eps1_def}) requires that
the fitness of the resulting genotype~$x'$  is not less than the
fitness of the parents with probability at least~$\varepsilon_0$,
for some constant~$\varepsilon_0>0$, i.e.
\begin{equation}\label{eps_cross}
\varepsilon_0 \le \Pr\Big(\max\{f({x}'),f({y}')\} \ge
\max\{f({x}),f({y})\}\Big)
\end{equation}
for any ${x},{y} \in {\mathcal X}$. This condition is also
fulfilled for the single-point crossover with $\varepsilon_0 =
1-p_{\rm c}$, if $p_{\rm c}<1$ is a constant. Besides that,
Condition~(\ref{eps_cross}) is satisfied with $\varepsilon_0 = 1$
for the optimized crossover operators, where the offspring is
computed as a solution to the optimal recombination problem.
Polynomial-time optimized crossover routines are known for Maximum
Clique~\cite{BN98}, Set Packing, Set Partition and some other
NPO~problems~\cite{ErKov14I,ErKov14II}.

\paragraph{Bitwise Mutation and $K$-Bounded Neighborhood Mappings}
Let $D({x},{y})$ denote the Hamming distance between ${x}$ and
${y}$.

\begin{definition}\label{k_bounded} {\rm \cite{AP95}}
Suppose $\Pi$ is an NP~optimization problem. A neighborhood
mapping~$\mathcal{N}$ is called $K$-bounded, if for any ${x} \in
{\rm \mbox{\rm Sol}}$ and ${y} \in \mathcal{N}({x})$ holds
${D({x},{y}) \le K}$, where $K$ is a constant.
\end{definition}

The bitwise mutation operator~$\mbox{Mut}^*$ outputs a
string~${y}$, given a string~${x}$, with probability $p_{\rm
m}^{D({x},{y})}(1-p_{\rm m})^{n-D({x},{y})}$. Note that
probability~$p_{\rm m}^j(1-p_{\rm m})^{n-j}$, as a function
of~$p_{\rm m}$, \ $p_{\rm m} \in [0,1]$, attains its minimum at
$p_{\rm m}=j/n$. The following proposition gives a lower bound for
the probability $\Pr({\mbox{Mut}^*({x})={y}})$, which is valid for
any ${y} \in \mathcal{N}({x})$, assuming that~$p_{\rm m}=K/n$.

\begin{proposition}\label{optimal_bound}
Suppose the neighborhood mapping~$\mathcal{N}$ is $K$-bounded,
$K\le n/2$ and $p_{\rm m}=K/n$. Then for any ${x} \in {\rm
\mbox{\rm Sol}}$ and any ${y} \in \mathcal{N}({x})$ holds
$$
\Pr(\mbox{\rm Mut}^*({x})={y}) \ge K^K/(en)^K.
$$
\end{proposition}

The proof may be found in the appendix.

\section{Expected First Hitting Time of Target Subset}
\label{sec:drift}

This section is based on the drift analysis of GAs
from~\cite{CDEL14}. Suppose that for some~$m$ there is an ordered
partition of~$\mathcal{X}$ into subsets $(A_1,\dots,A_{m+1})$
called {\em levels}. Level~$A_{m+1}$ will be the target level in
the subsequent analysis. The target level may be chosen as the set
of solutions with maximal fitness or the set of local optima or
the set of $\rho$-approximation solutions for some approximation
factor~$\rho
> 1$. A well-known example of partition is the \emph{canonical}
partition, where each level regroups solutions having the same
fitness value (see e.g.~\cite{bib:l11}). For $j\in [m+1]$ we
denote by $H_j:=\cup_{i=j}^{m+1} A_i$, the union of all levels
starting from level~$j$.

Given a levels partition, there always exists a total
order~"$\succ$" on~$\mathcal X$, which is aligned with
$(A_1,\dots,A_{m+1})$ in the sense that $x \succ y$ for any $x\in
A_{j}$, $y\in A_{j-1}$, $j\in[m+1]$. W.l.o.g. in what follows the
elements of a population vector~$P \in \mathcal X^{\lambda}$ will
be assumed to form a non-increasing sequence $x^{1}, x^{2}, \dots
,x^{\lambda}$ in terms of~"$\succ$" order: $x^{1}\succeq
x^{2}\succeq  \dots \succeq x^{\lambda}$. For any
constant~$\gamma\in(0,1)$, the individual~$x^{\lceil \gamma
\lambda\rceil}$ will be referred to as the $\gamma$-ranked
individual of the population.

The \emph{selective pressure} of a selection mechanism $\rm Sel$
is defined as follows. For any $\gamma \in (0,1)$ and population
$P$ of size $\lambda$, let $\beta(\gamma,P)$ be the probability of
selecting an individual from $P$ that belongs to the same or
higher level as the individual with
rank~$\lceil\gamma\lambda\rceil$, i.e.
$$
\beta(\gamma, P):= \sum_{i: x^i \in H_{j(\gamma)}} p_{sel} (i \mid
P),
$$
where $j(\gamma)$ is such that $x^{\lceil \gamma \lambda\rceil}
\in A_{j(\gamma)}$.



\begin{theorem}\label{GA_bitwise_mut} Given a  partition
  $(A_1,\ldots,A_{m+1})$ of $\mathcal{X}$, let $T := \min\{t\lambda
  \mid |P_t\cap A_{m+1}|>0\}$ be the runtime of ${\mathcal GA}$.
  If there exist parameters\linebreak
  $s_1,\dots,s_m,s_*,p_0,\varepsilon_1\in(0,1]$,
  $\delta>0$, and a constant $\gamma_0 \in (0,1)$ such that for all
  $j\in[m]$, and $\gamma \in (0,\gamma_0)$
  \begin{description}[noitemsep]
  \item[(C1)] $\displaystyle \pmut(y\in H_{j+1} \mid x\in H_{j}) \geq s_j \geq
  s_*$,
  \item[(C2)] $\displaystyle \pmut(y\in H_{j+1}\mid x\in H_{j+1})\geq
  p_0$,
  \item[(C3)] $\displaystyle \pxor(x\in H_{j+1}\mid u\in H_{j}, v\in H_{j+1})\geq
  \varepsilon,$
  \item[(C4)] $\beta(\gamma,P)\geq
  \gamma\sqrt{\frac{1+\delta}{p_0\varepsilon\gamma_0}}$ for any $P\in \mathcal{X}^\lambda$
  \item[(C5)] $\displaystyle \lambda \geq
    \frac{2}{a}\ln\left(\frac{32mp_0}{(\delta\gamma_0)^2cs_*\psi}\right)$
    with $\displaystyle a := \frac{\delta^2 \gamma_0}{2(1+\delta)}$,
    $\psi := \min\{\frac{\delta}{2},\frac{1}{2}\}$
    and $c := \frac{\psi^4}{24}$
  \end{description}
  then
  $\expect{T} \leq \frac{2}{c\psi}\left(m\lambda(1+\ln(1+c\lambda)) +\frac{p_0}{(1+\delta)\gamma_0}\sum_{j=1}^{m}\frac{1}{s_j}\right)$.
\end{theorem}

Informally, condition~(C1) requires that for each element of
subset~$H_j$, there is a lower limit~$s_j$ on probability to
mutate it into level~$j+1$ or higher. Condition~(C2) requires that
there exists a lower limit~$p_0$ on the probability that the
mutation will not "downgrade" an individual to a lower level.
Condition~(C3) follows from lower bound~(\ref{eps_cross}) assuming
$\varepsilon:=\varepsilon_0$ or from lower
bound~(\ref{eqn:eps1_def}) with $\varepsilon:=\varepsilon_1$ in
the case of the canonical partition. Condition~(C4) requires that
the selective pressure induced by the selection mechanism is
sufficiently strong. Condition~(C5) requires that the population
size~$\lambda$ is sufficiently large.

Unfortunately, Conditions (C3) and (C4) are unlikely to be
satisfied when the target subset~$A_{m+1}$ contains some less fit
solutions than the solutions from level~$A_{m}$, e.g.
when~$A_{m+1}$ is the set of all local optima. In order to adapt
Theorem~\ref{GA_bitwise_mut} to analysis of such situations we
first prove the following corollary with relaxed version of
conditions (C3),(C4) and a slightly strengthened version of~(C2).

\begin{corollary}\label{GA_for_local_search} Given a  partition
  $(A_1,\ldots,A_{m+1})$ of $\mathcal{X}$, let $T := \min\{t\lambda
  \mid |P_t\cap A_{m+1}|>0\}$ be the runtime of ${\mathcal GA}$.
  If there exist parameters\linebreak
  $s_1,\dots,s_m, s_*, p_0, \varepsilon\in(0,1]$,
  $\delta>0$, and a constant $\gamma_0 \in (0,1)$ such that for all
  $\gamma \in (0,\gamma_0)$
  \begin{description}[noitemsep]
  \item[(C1)] $\displaystyle \pmut(y\in H_{j+1} \mid x\in H_{j}) \geq s_j \geq
  s_*$, $j\in[m]$,
  \item[(C2')] $\displaystyle \pmut(x\mid x)\geq
  p_0$, $x \in {\mathcal X}$,
  \item[(C3')] $\displaystyle \pxor(x\in H_{j+1}\mid u\in H_{j}, v\in H_{j+1})\geq
  \varepsilon,$ $j\in[m-1]$,
  \item[(C4')] $\beta(\gamma,P)\geq
  \gamma\sqrt{\frac{1+\delta}{p_0\varepsilon\gamma_0}}$ for any $P\in (\mathcal{X}\backslash
  A_{m+1})^\lambda$,
  \item[(C5)] $\displaystyle \lambda \geq
    \frac{2}{a}\ln\left(\frac{32mp_0}{(\delta\gamma_0)^2cs_*\psi}\right)$
    with $\displaystyle a := \frac{\delta^2 \gamma_0}{2(1+\delta)}$,
    $\psi := \min\{\frac{\delta}{2},\frac{1}{2}\}$
    and $c := \frac{\psi^4}{24}$
  \end{description}
  then
  $\expect{T} \leq \frac{2}{c\psi}\left(m\lambda(1+\ln(1+c\lambda)) +\frac{p_0}{(1+\delta)\gamma_0}\sum_{j=1}^{m}\frac{1}{s_j}\right)$.
\end{corollary}

{\bf Proof.} Given a genetic algorithm ${\mathcal GA}$ with
certain initialization procedure for~$P^0$, selection
operator~${\rm Sel}$, crossover and mutation $\rm Cross$ and $\rm
Mut$ and population size~$\lambda$, consider a genetic algorithm
${\mathcal GA}'$ defined as the following modification of
${\mathcal GA}$.

\begin{itemize}
\item Let the initialization procedure for population~$P^0$ in ${\mathcal GA}'$
coincide with that of ${\mathcal GA}$.

\item Operator of selection ${\rm Sel}'(P)$  performs identically to operator~${\rm
Sel}(P)$, except for the cases when the input population~$P$
contains an element from~$A_{m+1}$. In the latter cases $\rm
Sel'(P)$ returns the index of the first representative
of~$A_{m+1}$ in~$P$.

\item Operator of crossover ${\rm Cross'}$ performs identically to $\rm Cross$
except for the cases when the input contains an element
from~$A_{m+1}$. In the latter cases an element of~$A_{m+1}$ is
just copied to the output of the operator.

\item Operator of mutation ${\rm Mut'}$ is the same as ${\rm Mut}$.

\item The population size in ${\mathcal GA}'$ is~$\lambda$.

\end{itemize}

Note that ${\mathcal GA}'$ meets Conditions~(C1)-(C5) of
Theorem~\ref{GA_bitwise_mut}. Indeed, Condition~(C2) follows
from~(C2'). Condition~(C3) is satisfied for $j\in[m-1]$ by (C3'),
and for $j=m$ it holds with $\varepsilon=1$ by definition of
operator $\rm Cross'$. Condition~(C4) is satisfied for any $P\in
(\mathcal{X}\backslash A_{m+1})^\lambda$ by (C4'), and in the
cases when population~$P$ contains at least one element
from~$A_{m+1}$, holds $\beta(\gamma,P)=1$ by definition of
operator $\rm Sel'$. Thus, by Theorem~\ref{GA_bitwise_mut},
$$
\expect{T'} \leq \frac{2}{c\psi}\left(m\lambda(1+\ln(1+c\lambda))
+\frac{p_0}{(1+\delta)\gamma_0}\sum_{j=1}^{m}\frac{1}{s_j}\right),
$$
where $T' := \min\{t\lambda \mid |P'_t\cap A_{m+1}|>0\}$ is
defined for the sequence of populations $P'_0,P'_1,\dots$ of
${\mathcal GA}'$.

Executions of ${\mathcal GA}$ and ${\mathcal GA}'$ before
iteration $T'/\lambda$ are identical. On iteration~$T'/\lambda$
both algorithms place elements of~$A_{m+1}$ into the population
for the first time. Thus, realizations of random variables $T'$
and $T$ coincide and $\expect{T}=\expect{T'}$.
 \qed

\section{Lower Bounds on Cumulative Selection\\ Probability}
\label{sec:beta_lower_bounds}

Let us see how to parameterise three standard selection mechanisms
in order to ensure that the selective pressure is sufficiently
high. We consider three selection operators with the following
mechanisms.

By definition, in \emph{$k$-tournament selection}, $k$ individuals
are sampled uniformly at random with replacement from the
population, and a fittest of these individuals is returned. In
$(\mu,\lambda)$-\emph{selection}, parents are sampled uniformly at
random among the fittest $\mu$ individuals in the population. The
ties in terms of fitness function are resolved arbitrarily.

A function $\alpha:\mathbb{R}\rightarrow\mathbb{R}$ is a ranking
function \cite{bib:GoldbergDeb} if $\alpha(x)\geq 0$ for all
$x\in[0,1]$, and $\int_0^1\alpha(x) dx = 1$. In ranking selection
with ranking function $\alpha$, the probability of selecting
individuals ranked $\gamma$ or better is
$\int_0^\gamma\alpha(x)dx$. We define \emph{exponential ranking}
parameterised by $\eta>0$ as $\alpha(\gamma):=\eta e^{\eta(1 -
  \gamma)}/(e^\eta - 1)$.

The following lemma is analogous to Lemma~1 from~\cite{CDEL14}.

\begin{lemma}\label{lemma:selection} \cite{CDEL14}
  Let the levels $A_1,\dots,A_m$ satisfy
\begin{equation}\label{eqn:relaxed_monoton}
f(x)<f(y) \ \mbox{for any} \ x\in A_{j-1}, \ y \in A_j, \
j=2,\dots,m.
\end{equation}
for all $x,y$ from $A_2,\dots,A_m$.

  Then for any constants $\delta'>0,$ $p_0 \in (0,1)$
  and $\varepsilon \in (0,1)$,
  there exist two constants $\gamma_0>0$ and $\delta>0$ such that
  \begin{enumerate}
  \item
$k$-tournament selection with $k \geq 4(1+\delta')/(\varepsilon
p_0)$,
\item
$(\mu,\lambda)$-selection with $\lambda/\mu \geq
  (1+\delta')/(\varepsilon p_0)$ and
\item exponential ranking selection with $\eta \geq
  4(1+\delta')/(\varepsilon p_0)$
  \end{enumerate}
satisfy~(C4'), i.e. $\beta(\gamma',P)\geq \gamma'\sqrt{
\frac{1+\delta}{p_0 \varepsilon \gamma'}}$ for any $\gamma' \in
(0,\gamma_0]$ and any\linebreak ${P\in (\mathcal{X}\backslash
A_{m+1})^\lambda}$.
\end{lemma}

Note that the assumption of montonicity of mutation w.r.t. all
fitness levels (see~\cite{CDEL14}) is substituted here by
Inequality~(\ref{eqn:relaxed_monoton}).

{\bf Proof.} Denote $\varepsilon':=\varepsilon p_0$.

1. Consider $k$-tournament selection. In order to select an
individual from the same level as the $\gamma$-ranked individual
or higher, by Inequality~(\ref{eqn:relaxed_monoton}) it is
sufficient that the randomly sampled tournament contains at least
one individual with rank~$\gamma$ or higher. Hence, one obtains
for $0 <\gamma < 1,$
$$
\beta(\gamma) > 1 - (1 - \gamma)^k.
$$
Note that
$$
(1 -\gamma)^k < e^{-\gamma k} < \frac{1}{1+\gamma k}.
$$

So for $k \geq 4(1+\delta')/\varepsilon'$, we have
\begin{align*}
  \beta(\gamma)
    &\geq 1 - \frac{1}{1+\gamma k}
     \geq 1 - \frac{1}{1+4\gamma(1+\delta')/\varepsilon'}
     = \frac{4\gamma(1+\delta')/\varepsilon'}{1+4\gamma(1+\delta')/\varepsilon'}
\end{align*}
If $\gamma_0 := \varepsilon'/(4(1+\delta'))$, then for all
$\gamma' \in (0, \gamma_0]$ it holds that
$4(1+\delta')/\varepsilon' \leq 1/\gamma'$ and
\begin{align*}
  \beta(\gamma')
    &\geq \frac{\gamma' 4(1+\delta')/\varepsilon'}{\gamma' (1/\gamma') + 1}
     = \frac{2(1+\delta')\gamma'}{\varepsilon'}
\end{align*}
\begin{align*}
     = \sqrt{\frac{(1+\delta')}{\varepsilon' (\varepsilon'/4(1+\delta'))}}
     \gamma'
     = \sqrt{\frac{(1+\delta')}{\varepsilon' \gamma_0}} \gamma'
\end{align*}

2. In $(\mu,\lambda)$-selection, for all $\gamma \in (0,
\mu/\lambda]$ we have $\beta(\gamma) = \lambda\gamma/\mu$ if
$\gamma \lambda \le \mu$, and $\beta(\gamma) = 1$ otherwise (see
by Inequality~(\ref{eqn:relaxed_monoton})). It suffices to pick
$\gamma_0 := \mu/\lambda$ so that with $\lambda/\mu \geq
(1+\delta')/\varepsilon'$, for all $\gamma' \in (0, \gamma_0]$.
Then
\begin{align*}
  \beta(\gamma')
    & \geq \frac{\lambda \gamma'}{\mu}
    = \sqrt\frac{\lambda^2}{\mu^2}\gamma'
    = \sqrt\frac{\lambda}{\mu\gamma_0}\gamma'
    \geq \sqrt\frac{1+\delta'}{\varepsilon'\gamma_0}\gamma'.
\end{align*}

3. In exponential ranking selection, we have
\begin{align*}
  \beta(\gamma)
    \geq \int_{0}^\gamma \frac{\eta e^{\eta(1 - x)}dx}{e^\eta - 1}
    = \left(\frac{e^\eta}{e^{\eta}-1}\right)\left(1 - \frac{1}{e^{\eta\gamma}}\right)
    \geq 1 - \frac{1}{1 + \eta\gamma}
\end{align*}
The rest of the proof is similar to tournament selection with
$\eta$ in place of $k$, e.g. based on the input condition on
$\eta$, it suffices to pick $\gamma_0 :=
\varepsilon'/(4(1+\delta'))$. \qed

\section{Expected First Hitting Time of the Set of Local Optima}

Suppose an NP~maximization problem~$\Pi=({\rm Inst},\mbox{\rm
Sol},F_I)$ is given and a neighborhood mapping~$\mathcal{N}_I$ is
defined. Given $I \in {\rm Inst}$, let~$s(I)$ be a lower bound on
the probability that a mutation operator transforms a given
feasible solution~$\x$ into a specific neighbor~${y} \in {\mathcal
N}_I ({x})$, i.e.
\begin{equation}\label{eqn:s_def}
s(I) \le \min_{{x} \in {\rm Sol}(I), \ {y} \in {\mathcal N}_I
({x})}{\Pr}(\mbox{Mut}({x})={y}).
\end{equation}
The greater the value~$s(I)$, the more consistent is the mutation
with the neighborhood mapping~${\mathcal N}_I$.

In Subsections \ref{subsec:noinfeas} and \ref{subsec:illustrate},
the symbol~$I$ is suppressed in the notation for brevity. The size
of population~$\lambda$ and the selection parameters~$k,\mu$ and
$\eta$, the number of levels~$m$ and the fitness function~$f$ are
supposed to depend on the input data~$I$ implicitly.

The set of all local optima is denoted by~$\mathcal{LO}$ (note
that global optima also belong to~$\mathcal{LO}$).

\subsection{No Infeasible Solutions} \label{subsec:noinfeas}
In many well-known NP~optimization problems, such as the Maximum
Satisfiability Problem~\cite{GJ}, the Maximum Cut
Problem~\cite{GJ} and the Ising Spin Glass Model~\cite{Barahon},
the set of feasible solutions is the whole search space, i.e.
${\rm Sol}=\{0,1\}^{n}=\mathcal X$. Let us consider the GAs
applied to problems with such a property.

We choose $m$ to be the number of fitness values $f_1<\dots<f_{m}$
attained by the solutions from~$\mathcal{X} \backslash
\mathcal{LO}$. Then starting from any point, the local search
method finds a local optimum within at most~$m$ steps. Let us use
a modification of canonical $f$-based partition where all local
optima are merged together:
\begin{equation} \label{eqn:partit_1}
A_j := \{x\in \mathcal{X} | f(x) = f_j\} \backslash \mathcal{LO},
\ j\in [m],
\end{equation}
\begin{equation} \label{eqn:partit_2}
A_{m+1} := \mathcal{LO}.
\end{equation}
Application of Corollary~\ref{GA_for_local_search} and
Lemma~\ref{lemma:selection} w.r.t. this partition yields the
following theorem.

\begin{theorem}\label{theorem_GA_LS1}
Suppose that
\begin{itemize}
\item $\pmut(y \mid x)\ge s$ for any $\ x \in {\rm Sol}, \ y \in
{\mathcal N} (x),$
\item Conditions~(C2') and (C3') are satisfied for
some constants~$p_0>0$ and $\varepsilon>0$,
\item ${\rm Sol}={\mathcal X}$,
\item $\mathcal GA$ is using either
$k$-tournament selection with
$k\ge\frac{4(1+\delta')}{\varepsilon_0 p_0}$, or
 $(\mu,\lambda)$-selection with
  $\frac{\lambda}{\mu} \ge \frac{(1+\delta')}{\varepsilon_0 p_0}$
   or exponential ranking selection with
   $\eta\ge\frac{4(1+\delta')}{\varepsilon_0 p_0}$ for some constant~$\delta'>0$.
\end{itemize}
Then there exist two constants~$b$ and $b'$, such that for
population size~$\lambda \ge b\ln\left(\frac{m}{s}\right)$, a
local optimum is reached for the first time after at most
$b'(m\lambda \ln \lambda + \frac{m}{s})$ fitness evaluations on
average.
\end{theorem}

A similar result for the $\mathcal GA$ with tournament selection
and two-offspring crossover was obtained
in~\cite{Er11,bib:erarxiv13} without a drift analysis. In
particular, Lemma~1 and Proposition~1 in~\cite{bib:erarxiv13}
imply that with appropriate settings of parameters, a non-elitist
genetic algorithm reaches a local optimum for the first time
within $O\left(\frac{m\ln m}{s}\right)$ fitness evaluations on
average. The upper bound from Theorem~\ref{theorem_GA_LS1} in the
present paper has advantage in to the bound
from~\cite{bib:erarxiv13} if $1/s$ is at least linear in~$m$.
(Note that the size of many well-known neighborhoods grows as some
polynomial of~$m$.)

\subsection{Illustrative Examples} \label{subsec:illustrate}

\paragraph{Royal Road Functions} Let us consider a
family of {\em Royal Road Functions}~$RR_{n,r}$ defined on the
basis of the principles proposed by M.~Mitchell, S.~Forrest, and
J.~Holland in~\cite{MFH92}. The function~$RR_{n,r}$ is defined on
the search space ${\mathcal X}=\{0, 1\}^n$, where $n$ is a
multiple of~$r$, and the set of indices~$[n]$ is partitioned into
consecutive blocks of~$r$ elements each. By definition
$RR_{n,r}(x)$ is the number of blocks where all bits are equal
to~1.

We consider a crossover operator, denoted by ${\rm Cross}^{p_c}$,
which returns one of the parents unchanged with probability
$1-p_c$. In particular, ${\rm Cross}^{p_c}$ may be built up from
any standard crossover operator so that with probability $p_c$ the
standard operator is applied and the offspring is returned,
otherwise with probability $1-p_c$ one of the two parents is
returned with equal probabilities.

The following corollary for royal road functions~$RR_{n,r}(x)$
results from Theorem~\ref{theorem_GA_LS1} with the neighborhood
defined by the Hamming distance with radius $r$.

\begin{corollary} \label{corr:royal}
Suppose that the ${\mathcal GA}$ uses a ${\rm Cross}^{p_c}$
 crossover operator with $p_c$ being any
constant in $[0,1)$, the bitwise mutation with mutation rate
$p_{\rm m}=\chi/n$ for a constant $\chi>0$, $k$-tournament
selection with $k\geq 8(1+\delta)e^{\chi}/(1-p_c)$, or
$(\mu,\lambda)$-selection with $\lambda/\mu\geq
2(1+\delta)e^{\chi}/(1-p_c)$, or exponential ranking selection
with $\eta\geq 8(1+\delta)e^{\chi}/(1-p_c)$, where $\delta>0$ is a
constant. Then there exists a constant $b>0$ such that the GA with
population size~$\lambda = \lceil b \ln n \rceil$, has expected
runtime $O(n^{r+1})$ on~$RR_{n,r}(x)$.
\end{corollary}
{\bf Proof.}

Note that fitness function~$RR_{n,r}(x)$ of any solution~$x$ with
some non-optimal bits (i.e. bits equal to zero) can be increased
by an improving move within Hamming neighborhood of radius~$r$. So
there is just one local optimum and it is the global
optimum~$x=(1,\dots,1)$. We now apply Theorem~\ref{theorem_GA_LS1}
with the canonical partition $A_{j} : = \{x \mid RR_{n,r}(x) =
j-1\}$ for
  $j\in [m+1]$ where $m:= {n}/{r}.$

The probability of not flipping any bit position by mutation is
$$
(1-\chi/n)^n=(1-\chi/n)^{(n/\chi-1)\chi}(1-\chi/n)^\chi
      \geq e^{-\chi}(1-\chi/n)^\chi.
$$
In the rest of the proof we assume that~$n$ is sufficiently large
to ensure that $(1-\chi/n)^n\ge \frac{e^{-\chi}}{1+\delta/2}$ for
the constant~$\delta>0$. Let $p_0:=\frac{e^{-\chi}}{1+\delta/2}.$

The lower bound to upgrade probability may be found if we consider
the worst-case scenario where only one block contains some
incorrect bits and the number of such bits is~$r$. Then $s:=
(\chi/n)^r(1-\chi/n)^{n-r} = \Omega(1/n^r)$.

  We can put $\varepsilon_0 := (1-p_c)/2$ because the crossover operator
  returns one of the parents unchanged with probability $1-p_c$, and with
  probability at least $1/2$, this parent is not less fit than the other one.
  Then conditions of Theorem~\ref{theorem_GA_LS1} regarding
  $k, \mu, \lambda$, and $\eta$ are satisfied
  for the constant $\delta':=\frac{1+\delta}{1+\delta/2}-1>0$.

  It therefore follows by Theorem~\ref{theorem_GA_LS1} that
  there exists a constant $b>0$ such that if the population size is
  $\lambda=\lceil b\ln n \rceil$, the expected runtime of $\mathcal GA$ on~$RR_{n,r}(x)$ is
  upper bounded by $b'(n(\lambda \ln \lambda + n^r))$ for some constant~$b'$.
  \qed

The corollary implies that the $\mathcal GA$ with proper
population size has a polynomial runtime on the royal road
functions if~$r$ is a constant.

\paragraph{Vertex Cover Problems with Regular Structure}
In general, given a graph $G=(V,E)$, the Vertex Cover
Problem~(VCP) asks for a subset $C \subset V$ (called a {\it
vertex cover}), such that every edge $e\in E$ has at least one
endpoint in~$C$. The size of~$C$ should be minimized. Let us
consider a representation of the problem solutions, where $n=|E|$,
${\mathcal X}=\{0,1\}^{n}$ and each coordinate $x^i\in\{0,1\},
i=1,\dots,|E|$ of $x$ corresponds to an edge $e_i\in E$, assigning
one of its endpoints to be included into the cover $C(x)$ (one
endpoint of $e_i$ is assigned if $x^i=0$ and the other one is if
$x^i=1$). Thus, $C(x)$ contains all vertices, assigned by at least
one of the coordinates of $x$, and the feasibility of $C(x)$ is
guaranteed. This representation is a special case of the so-called
{\it non-binary representation} for a more general set covering
problem (see e.g.~\cite{BeCh96}). The fitness function is by
definition $f(x):=|V|-|C(x)|$.

Family of vertex covering instances $G(\kappa), \kappa=1,2,\dots$
consists of VCP problems with ${n}=3\kappa$ where~$G$ is a union
of $\kappa$~disjoined cliques of size~3 (triangles). An optimal
solution contains a couple of vertices from each clique and there
are $3^{\kappa}$ optimal solutions.

Consider the neighborhood system defined by Hamming distance with
radius~1. All local optima for $G(\kappa)$ are globally optimal in
this case. For the bit-wise mutation operator with mutation rate
$p_{\rm m}=1/n$ by Proposition~\ref{optimal_bound} we have $P({\rm
Mut}^*(x)=y)\ge 1/(en)$ for any $y\in {\mathcal N}(x), \
x\in{\mathcal X}$. Analogously to Corollary~\ref{corr:royal} we
obtain

\begin{corollary} \label{corr:G}
Suppose that the ${\mathcal GA}$ uses a ${\rm Cross}^{p_c}$
crossover operator with $p_c$ being any constant in $[0,1)$, the
bitwise mutation with mutation rate $p_{\rm m}=1/n$,
$k$-tournament selection with $k\geq 8e(1+\delta)/(1-p_c)$, or
$(\mu,\lambda)$-selection with $\lambda/\mu\geq
2e(1+\delta)/(1-p_c)$, or exponential ranking selection with
$\eta\geq 8e(1+\delta)/(1-p_c)$, where $\delta>0$ is a constant.
Then there exists a constant $b>0$, such that the GA with
population size~$\lambda= b \lceil\ln n\rceil$, has expected
runtime~$O(n \ln n \ln \ln n)$ on the VCP family~$G(\kappa)$.
\end{corollary}

It is interesting that the VCP instances $G(\kappa)$ in integer
linear programming formulation are hard for the Land and Doig
branch and bound algorithm (for a description of Land and Doig
algorithm see e.g.~\cite{Schrijver}, Chapt.~24). This exact
algorithm makes $2^{\kappa+1}$ branchings on the problems from
$G(\kappa)$ (see~\cite{Saiko}).

\subsection{The General Case of NP Optimization\\ Problems}

Consider the general case where ${\rm Sol}(I)$ may be a proper
subset of~${\mathcal X}$. Let us add another modification to the
levels partition. Besides merging all local optima we assume that
all infeasible solutions constitute level~$A_1$. The rest of
solutions are stratified by their objective function values.
Let~$m-1$ be the number of fitness values $f_2<\dots<f_{m}$
attained by the feasible solutions from~${\rm Sol}(I) \backslash
\mathcal{LO}_I$.
\begin{equation} \label{eqn:partit_1a}
 A_1 := \mathcal{X} \backslash {\rm Sol}(I),
\end{equation}
\begin{equation} \label{eqn:partit_2a}
 A_j := \{x\in {\rm Sol}(I) |
f(x) = f_{j}\} \backslash \mathcal{LO}_I, \ j=2,\dots,m,
\end{equation}
\begin{equation} \label{eqn:partit_3a}
A_{m+1} := \mathcal{LO}_I.
\end{equation}
Application of Corollary~\ref{GA_for_local_search} with this
partition yields the following lemma.

\begin{lemma}\label{lemma_GA_LS2}
Suppose that Condition~(C2') holds and
  \begin{description}[noitemsep]
  \item[(L1)] $\sigma(I) \le \min\{\Pr(\mbox{Mut}({x})={y}) \mid {x} \in {\rm Sol}(I), \ {y} \in {\mathcal N}_I
({x})\}$.
  \item[(L2)] $\sigma(I) \le \Pr(\mbox{Mut}({x}) \in {\rm Sol}(I))$, \ $x\in {\mathcal
  X}\backslash {\rm Sol(I)}$.
  \item[(L3)] Inequality~(\ref{eps_cross})
  holds for some positive constant~${\varepsilon_0}$
  \end{description}
and $\mathcal GA$ is using either $k$-tournament selection with $k
\ge \frac{4(1+\delta')}{\varepsilon p_0}$ or
 $(\mu,\lambda)$-selection with
  $\frac{\lambda}{\mu}\ge \frac{(1+\delta')}{\varepsilon p_0}$
   or exponential ranking selection with $\eta
\ge \frac{4(1+\delta')}{\varepsilon p_0}$ for some
constant~$\delta'>0$.

Then there exist two constants~$b$, and $b'$ such that for
population size~$\lambda \ge
b\ln\left(\frac{m(I)}{\sigma(I)}\right)$, a local optimum is
reached for the first time after at most\linebreak
$b'\left(m(I)\lambda \ln \lambda+ \frac{m(I)}{\sigma(I)}\right)$
fitness evaluations on average.
\end{lemma}

{\bf Proof.} Assumption~(L1) is equivalent to
Inequality~(\ref{eqn:s_def}) with $\sigma(I)\equiv s(I)$.
Condition~(L2) imposes a lower bound on probability of producing
feasible solutions by mutation of an infeasible bitstring. Thus
together (L1) and (L2) give the lower bound for~(C1).
Condition~(L3) implies (C3'). \qed

Operators~$\mbox{Mut}$ and $\mbox{Cross}$ are supposed to be
efficiently computable and the selection procedure requires only a
polynomial time. Therefore the time complexity of computing a pair
of offspring in the $\mathcal GA$ is polynomially bounded and the
following theorem holds.

\begin{theorem}\label{GA_LS}
If problem~$\Pi=({\rm Inst},\mbox{\rm Sol}(I),F_I)$ is
polynomially bounded, Conditions~(C2'), (L1), (L2) and~(L3) are
satisfied for a lower bound~$\sigma(I),$  $1/\sigma(I) \in {\rm
Poly},$ and positive constants~$\varepsilon_0>0$ and $p_0>0$,
then~$\mathcal GA$ using tournament selection or
$(\mu,\lambda)$-selection or exponential ranking selection with a
suitable choice of parameters first visits a local optimum on
average in polynomially bounded time.
\end{theorem}


Note that Condition~(L2) in formulation of Theorem~\ref{GA_LS} can
not be dismissed. Indeed, suppose that problem~$\Pi=({\rm
Inst},\mbox{\rm Sol}(I),F_I)$ is polynomially bounded,
and consider a~$\mathcal GA$ where the mutation operator has the
following properties. On one hand~${\rm Mut}$ never outputs a
feasible offspring, given an infeasible input. On the other hand,
given a feasible genotype~$\x$, ${\rm Mut}(\x)$ is infeasible with
a {\em positive} probability~$\pi(\x,I)$,
$0<\epsilon(I)<\pi(\x,I)$. Finally assume that the initialization
procedure produces no local optima in population~$P^0$. Now all
conditions of Theorem~\ref{GA_LS} can be satisfied, but with a
positive probability of at least~$\epsilon(I)^{\lambda}$ the whole
population~$P^1$ consists of infeasible solutions, and subject to
this event all populations~$P^1,P^2,\dots$ are infeasible.
Therefore, expected hitting time of a local optimum is unbounded.

In order to estimate applicability of Theorem~\ref{GA_LS} it is
sufficient to recall that the set $\mathcal{N}({x})$ for $x\in
{\rm Sol}(I)$ may be enumerated efficiently by definition, so
there exists a mutation operator~$\mbox{Mut}(\x)$ that generates a
uniform distribution over~$\mathcal{N}({x})$
if~$\mathcal{N}({x})\ne \emptyset$ and every point
in~$\mathcal{N}({x})$ is selected with probability at
least~$\sigma(I), \ 1/\sigma(\cdot) \in{\rm Poly}$. To deal with
the cases where $x\not\in {\rm Sol}(I)$ or $x\in {\rm Sol}(I)$ but
$\mathcal{N}({x})=\emptyset$, we can recall that there are large
classes of NP-optimization problems, where at least one feasible
solution~$y_I$ is computable in polynomial time (see e.g. the
classes PLS in~\cite{JohPapYan88} and GLO in~\cite{AP95}). For
such problems in case of $x\not\in {\rm Sol}(I)$ or
$\mathcal{N}({x})=\emptyset$, a mutation operator may output the
feasible solution~$y_I$ with probability~1.

Alternatively we can consider a {\em repair heuristic} (see
e.g.~\cite{BeCh96}) which follows some standard mutation operator
and, if the output of mutation is infeasible then the heuristic
substitutes this output by a feasible solution, e.g.~$y_I$.

\section{Analysis of Guaranteed Local Optima Problems}
\label{sec:GLO}

In this section, Theorem~\ref{GA_LS} is used to estimate the GA
capacity of finding the solutions with approximation guarantee.

An algorithm for an NP~maximization problem  $\Pi =(\mbox{\rm
Inst},\mbox{\rm Sol}(I),F_I)$ has a {\em guaran\-teed
approxi\-mation ratio}~$\rho$, $\rho \ge 1$, if for any
instance~$I\in {\rm Inst}, \ {{\rm Sol}(I) \ne \emptyset,}$ it
delivers a feasible solution~$\x'$, such that
$$
F_I(\x') \ge \max\{F_I(\x) | \x \in {\rm Sol}(I)\}/\rho.
$$

In the case of an NP~minimization problem, the guaran\-teed
approxi\-mation ratio is defined similarly, except that the latter
inequality changes into
$$
F_I(\x') \le \rho \min\{F_I(\x) | \x \in {\rm Sol}(I)\}.
$$

\begin{definition}\label{GLO} {\rm \cite{AP95}}
A polynomially bounded NP~optimization problem~$\Pi$ belongs to
the class of {\em Guaranteed Local Optima}~(GLO) problems, if the
following two conditions hold:

1) At least one feasible solution~${y}_I \in {\rm \mbox{\rm Sol}}$
is efficiently computable for every instance~$I \in {\rm Inst}$;

2) A $K$-bounded neighborhood mapping~$\mathcal{N}_I$ exists, such
that for every instance~$I$, any local optimum of~$I$ with respect
to~$\mathcal{N}_I$ has a constant guaranteed approximation ratio.

\end{definition}

The class~{\rm GLO} contains such well-known {\rm NP}~optimization
problems as the Maximum Staisfiablity and the Maximum Cut
problems, besides that, on graphs with bounded vertex degree the
Independent Set problem, the Dominating Set problem and the Vertex
Cover problem also belong to GLO~\cite{AP95}.

If a problem~$\Pi$ belongs to GLO and $n$ is sufficiently large,
then in view of Proposition~\ref{optimal_bound}, for any ${x} \in
{\rm \mbox{\rm Sol}}$ and ${y} \in \mathcal{N}({x})$, the bitwise
mutation operator with~$p_{\rm m}=K/n$ satisfies the condition
$\Pr\{\mbox{Mut}^*({x})={y}\}^{-1} \in {\rm Poly}$. Besides that,
for a sufficiently large~$n$ for any $x\in {\rm \mbox{\rm Sol}}$
holds $\pmut(x \mid x)\geq e^{-K}/2=:p_0$, which is a constant.
Therefore, Theorem~\ref{GA_LS} implies the following

\begin{corollary}\label{GA_GLO}
If $\Pi\in {\rm GLO}$, then given suitable values of parameters,
$\mathcal GA$ with tournament selection or
$(\mu,\lambda)$-selection, a crossover operator that satisfies
Inequality~(\ref{eps_cross}) for some positive
constant~${\varepsilon_0}$ and the bitwise mutation followed by
repair heuristic, first visits a solution with a constant
guaranteed approximation ratio in polynomially bounded time on
average.
\end{corollary}


\section{\large{\textbf{Conclusion}}}

The obtained results indicate that if an NPO~problem is
polynomially bounded and a feasible solution is easy to find, then
a local optimum is computable in expected polynomial time by the
non-elitist GA with tournament selection or
$(\mu,\lambda)$-selection or exponential ranking selection.
Besides that, given suitable parameters, the non-elitist GA with
tournament selection or or $(\mu,\lambda)$-selection approximates
any problem from GLO class within a constant ratio in polynomial
time on average.

The paper does not take into account the possible {\em
improvement} of parent solutions in the crossover operator. The
further research might address the ways of using this potential.
Another open question is whether it is possible to prove an analog
of Theorem~\ref{GA_LS}, provided that the initial population
contains a sufficient amount of feasible solutions, and the
infeasible bitstrings mutate into feasible ones at least with
exponentially small probability.

\section{Acknowledgements} This
research received funding from the European Union Seventh
Framework Programme (FP7/2007-2013) under grant agreement no
618091 (SAGE), from EPSRC grant no~EP/F033214/1 (LANCS), and from
Russian Foundation for Basic Research grants~13-01-00862 and
15-01-00785. Early ideas where discussed at Dagstuhl Seminar 13271
Theory of Evolutionary Algorithms.

\bibliographystyle{splncs03}
\bibliography{bibliography}

\begin{thebibliography}{10}
\providecommand{\url}[1]{\texttt{#1}}
\providecommand{\urlprefix}{URL }

\bibitem{AartsLenstra}
Aarts, E.H.L., Lenstra, J.K.: Introduction. In: Aarts, E.H.L., Lenstra, J.K.
  (eds.) Local Search in Combinatorial Optimization. John Wiley \& Sons Ltd.
  New York, USA (1997)

\bibitem{ACGKMP}
Ausiello, G., Crescenzi, P., Gambosi, G., Kann, V., Marchetti-Spaccamela, A.,
  Protasi, M.: Complexity and approximation: combinatorial optimization
  problems and their approximability properties. Springer-Verlag, Berlin (1999)

\bibitem{AP95}
Ausiello, G., Protasi, M.: Local search, reducibility and approximability of
  $\mbox{NP}$-optimization problems. Information Processing Letters  54,
  73--79 (1995)

\bibitem{BN98}
Balas, E., Niehaus, W.: Optimized crossover-based genetic algorithms for the
  maximum cardinality and maximum weight clique problems. Journal of Heuristics
   4(2),  107--122 (1998)

\bibitem{Barahon}
Barahona, F.: On the computational complexity of $\mbox{Ising}$ spin glass
  models. Journal of Physics A, Mathematical and General  15,  3241--3253
  (1982)

\bibitem{BeCh96}
Beasley, J.E., Chu, P.C.: A genetic algorithm for the set covering problem.
  European Journal of Operation Research  94(2),  394--404 (1996)

\bibitem{BSW}
Beyer, H.G., Schwefel, H.P., Wegener, I.: How to analyse evolutionary
  algorithms. Theoretical Computer Science  287,  101--130 (2002)

\bibitem{CDEL14}
Corus, D., Dang, D.C., Eremeev, A.V., Lehre, P.K.: Level-based analysis of
  genetic algorithms and other search processes. In: Proceedings of Parallel
  Problem Solving from Nature (PPSN XIII). LNCS, vol. 8672, pp. 912--921.
  Springer (2014)

\bibitem{ErKov14I}
Eremeev, A., Kovalenko, J.: Optimal recombination in genetic algorithms for
  combinatorial optimization problems: Part~$\mbox{I}$. Yugoslav Journal of
  Operations Research  24(1),  1--20 (2014)

\bibitem{ErKov14II}
Eremeev, A., Kovalenko, J.: Optimal recombination in genetic algorithms for
  combinatorial optimization problems: Part~$\mbox{II}$. Yugoslav Journal of
  Operations Research  24(2),  165--186 (2014)

\bibitem{bib:erarxiv13}
Eremeev, A.V.: Non-elitist genetic algorithm as a local search method.
  Arxiv:1307.3463 (2013)

\bibitem{Er11}
Eremeev, A.: A genetic algorithm with tournament selection as a local search
  method. In: Proceedings of 15-th Baikal International School-Seminar
  ``Optimization methods and their applications''. pp. 94--99 (2011), in
  Russian

\bibitem{GJ}
Garey, M., Johnson, D.: Computers and intractability. A guide to the theory of
  NP-completeness. W.H. Freeman and Company (1979)

\bibitem{bib:GoldbergDeb}
Goldberg, D.E., Deb, K.: A comparative analysis of selection schemes used in
  genetic algorithms. In: Foundations of Genetic Algorithms. pp. 69--93. Morgan
  Kaufmann (1991)

\bibitem{bib:Gold89}
Goldberg, D.E.: Genetic Algorithms in search, optimization and machine
  learning. Addison-Wesley, MA, USA (1989)

\bibitem{Holl75}
Holland, J.: Adaptation in natural and artificial systems. University of
  Michigan Press (1975)

\bibitem{JohPapYan88}
Johnson, D.S., Papadimitriou, C.H., Yannakakis, M.: How easy is local search?
  Journal of computer and system sciences  37(1),  79--100 (1988)

\bibitem{Ko82}
Ko, K.: Some observations on the probabilistic algorithms and np-hard problems.
  Information Processing Letters  4,  39--43 (1982)

\bibitem{Krish}
Krishnakumar, K.: Micro-genetic algorithms for stationary and non-stationary
  function optimization. In: Proceedings of SPIE: Intelligent Control and
  Adaptive Systems. pp. 289--296 (1989)

\bibitem{bib:l11}
Lehre, P.K.: Fitness-levels for non-elitist populations. In: Proceedings of
  Genetic and Evolutionary Computation Conference (GECCO 2011). pp. 2075--2082
  (2011)

\bibitem{MFH92}
Mitchell, M., Forrest, S., Holland, J.: The royal road for genetic algorithms:
  Fitness landscapes and $\mbox{GA}$ performance. In: Varela, F., Bourgine, P.
  (eds.) Proceedings of the First European Conf. on Artificial Life. pp.
  245--254. MIT Press (1992)

\bibitem{RR02}
Reeves, C.R., Rowe, J.E.: Genetic algorithms: principles and perspectives.
  Kluwer Acad. Pbs. (2002)

\bibitem{Reeves}
Reeves, C.R.: Genetic algorithms for the operations researcher. INFORMS Journal
  on Computing  9(3),  231--250 (1997)

\bibitem{Saiko}
Saiko, L.A.: Studying the cardinality of $\mbox{L}$-coverings for some covering
  problems. In: Discrete Optimization and Analysis of Complex Systems, pp.
  76--97. Vychisl. Tsentr, Novosibirsk (1989), in Russian

\bibitem{Schrijver}
Schrijver, A.: Theory of linear and integer programming, vol.~2. John Wiley \&
  Sons (1986)

\bibitem{Vitanyi}
Vit\'{a}nyi, P.M.B.: A discipline of evolutionary programming. Theor. Comp.
  Sci.  24(1--2),  3--23 (2000)

\bibitem{Yank99}
Yankovskaya, A.: Test pattern recognition with the use of genetic algorithms.
  Pattern Recognition and Image Analysis  9(1),  121--123 (1999)

\end{thebibliography}

\section*{Appendix}

{\bf Proof of Proposition~\ref{optimal_bound}.} For any ${x} \in
{\rm \mbox{\rm Sol}}$ and ${y} \in \mathcal{N}({x})$ holds
$$
\Pr(\mbox{Mut}^*({x})={y})
=\left(\frac{K}{n}\right)^{D({x},{y})}\left(1-\frac{K}{n}\right)^{n-D({x},{y})}
$$
$$
\ge
 \left(\frac{K}{n}\right)^K
 \left(1-\frac{K}{n}\right)^{n-K},
$$
since $p_{\rm m} =K/n\le 1/2$. Now $\frac{\partial}{
\partial n} (1-K/n)^{n-K} <0$ for $n>K$, and besides that, $(1-K/n)^{n-K} \to 1/e^K$
as~$n\to \infty$. Therefore $(1-K/n)^{n-K} \ge 1/e^K$, which
implies the required inequality. \qed

\end{document}